\documentclass{article}
\usepackage{ijcai25}

\usepackage{times}
\usepackage{soul}
\usepackage{url}
\usepackage[hidelinks]{hyperref}
\usepackage[utf8]{inputenc}
\usepackage[small]{caption}
\usepackage{graphicx}
\usepackage{amsmath}
\usepackage{amssymb}
\usepackage{amsthm}
\usepackage{booktabs}
\usepackage{algorithm}
\usepackage{algorithmic}
\usepackage[switch]{lineno}
\usepackage{multirow}
\usepackage{float}

\title{Syllables to Scenes: Literary-Guided Free-Viewpoint 3D Scene Synthesis from Japanese Haiku}

\author{
Chunan Yu $^1$
\and
Yidong Han $^1$\and
Chaotao Ding $^{5}$\and
Ying Zang $^{1,*}$\and
Lanyun Zhu $^{6}$\\
Xinhao Chen $^{7}$\and
Zejian Li $^{3}$\and
Renjun Xu $^{4,*}$\And
Tianrun Chen $^{2,5,*}$\\
\affiliations
$^1$ School of Information Engineering, Huzhou University \\
$^2$ College of Computer Science and Technology, Zhejiang University \\
$^3$ College of Software Engineering, Zhejiang University \\
$^4$ Center for Data Science, Zhejiang University \\
$^5$ KOKONI 3D, Moxin (Huzhou) Technology Co., LTD. \\
$^6$ Singapore University of Technology and Design \\
$^7$ School of Humanities, Wenzhou University \\
\emails
02750@zjhu.edu.cn\and
rux@zju.edu.cn\and
tianrun.chen@kokoni3d.com
}

\begin{document}

\twocolumn[{
\renewcommand\twocolumn[1][]{#1}
\maketitle
\begin{center}
    \captionsetup{type=figure}
    \includegraphics[width=0.9\textwidth]{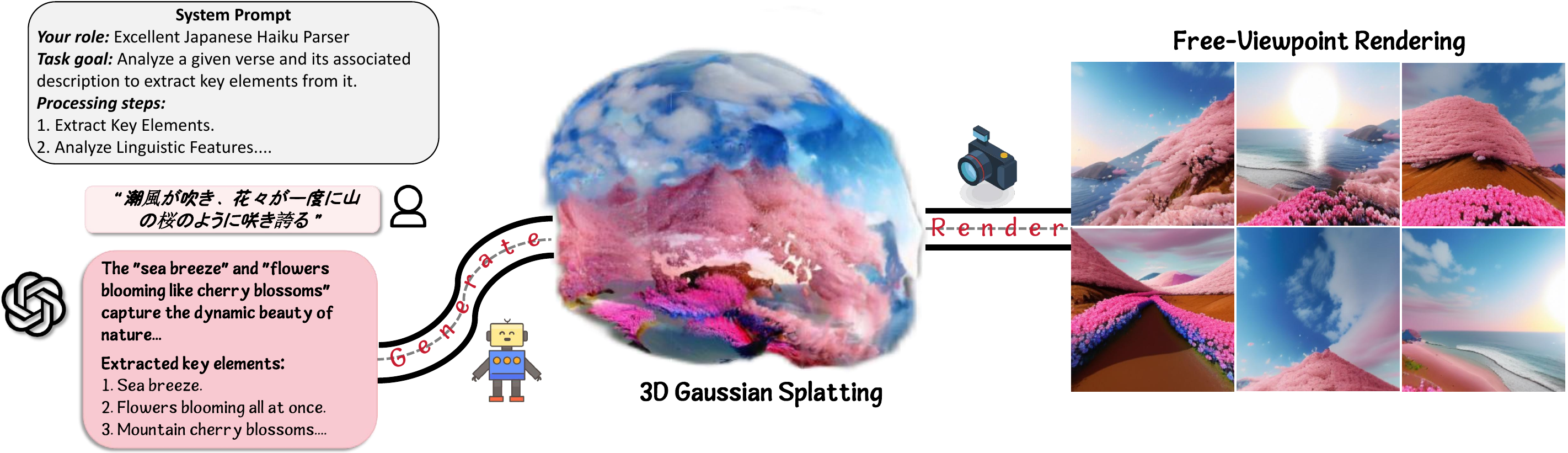}
    \caption{In this work, we propose a method to visualize traditional poetry in 3D. We parse classical Japanese Haiku using LLMs and transform the Haiku with its understanding and key element into coherent and high-quality 3D scenes represented by Gaussian Splatting. We show a paradigm of how to transform abstract concepts into explorable 3D environments, which also enables democratizing access to literary and artistic heritage. }
    \label{fig:1}  
\end{center}
}]

\begin{abstract}
    In the era of the metaverse, where immersive technologies redefine human experiences, translating abstract literary concepts into navigable 3D environments presents a fundamental challenge in preserving semantic and emotional fidelity. This research introduces HaikuVerse, a novel framework for transforming poetic abstraction into spatial representation, with Japanese Haiku serving as an ideal test case due to its sophisticated encapsulation of profound emotions and imagery within minimal text. While existing text-to-3D methods struggle with nuanced interpretations, we present a literary-guided approach that synergizes traditional poetry analysis with advanced generative technologies. Our framework centers on two key innovations: (1) Hierarchical Literary-Criticism Theory Grounded Parsing (H-LCTGP), which captures both explicit imagery and implicit emotional resonance through structured semantic decomposition, and (2) Progressive Dimensional Synthesis (PDS), a multi-stage pipeline that systematically transforms poetic elements into coherent 3D scenes through sequential diffusion processes, geometric optimization, and real-time enhancement. Extensive experiments demonstrate that HaikuVerse significantly outperforms conventional text-to-3D approaches in both literary fidelity and visual quality, establishing a new paradigm for preserving cultural heritage in immersive digital spaces. Project website at: 
\url{https://syllables-to-scenes.github.io/}
\end{abstract}

\section{Introduction}
    In the metaverse era, where immersive media increasingly shapes human experiences, the quest to translate abstract concepts into explorable 3D environments remains a grand challenge. Recently, significant advancements have been made in generating 3D scenes from textual or visual prompts \cite{zhou2025dreamscene360,chung2023luciddreamer,poole2022dreamfusion}. Yet, amidst these strides, a vital and profoundly human domain has been left behind: ancient poetry. Generative models, with their transformative potential, are ultimately tools to serve humanity—to bridge the gap between emotion and expression, art and technology. This raises an intriguing question: \textit{If ancient poets had access to today’s generative models, could their vivid imagery have been transformed into immersive, navigable experiences?} 
    
    Poetry—as a timeless form of emotional and cultural expression—deserves this attention, not only to honor its heritage but to explore how technology can serve as a bridge to deeper human connections \cite{heidegger1975poetry}. While some prior efforts have explored converting Chinese classical poetry into pictures \cite{jiang2024poetry2image,chen2024automatic,li2021paint4poem}, these static representations fail to meet the immersive and interactive expectations of AR/VR. Poetry, by its very essence, invites personal interpretation and emotional connection—qualities that demand more than a static frame. The challenge lies not just in visualizing poetic imagery but in creating navigable 3D scenes that preserve the richness of poetic expression, offering users the opportunity to explore these worlds with their own perspectives and emotions.
    
    We selected Japanese Haiku as a representative form of classical poetry for this exploration. Unlike Chinese classical poetry, often grand in scope and rich in descriptive detail, Haiku’s essence lies in its brevity and precision \cite{harr1975haiku,ueda1963basho}. With just seventeen syllables, a Haiku distills emotions, seasons, and fleeting moments into a compact form, challenging both cultural sensitivity and technological sophistication to interpret. This research focuses on Haiku to highlight its delicate artistry and demonstrate how generative models can bridge the abstract realm of poetry with the tangible, immersive possibilities of 3D technology, serving as a profound medium for human emotional expression.
    
    At the core of our endeavor lies a bold ambition: to translate the nuanced beauty of Haiku into fully navigable 3D scenes that capture its emotional depth and visual subtlety. Accomplishing this demanded overcoming two formidable challenges. The first and greatest challenge lies in the intrinsic nature of Haiku itself. Haiku’s extreme brevity encapsulates multiple layers of meaning and a rich interplay of emotions and imagery. Attempts to input Haiku into popular text-to-image models like SDXL \cite{podell2023sdxl} or DALL-E \cite{ramesh2021zero} often fail to produce satisfactory results because these models struggle with Haiku's conciseness and its multifaceted elements. They miss the mark in faithfully visualizing the depth and complexity of Haiku, leaving us far from achieving our goal of accurately rendering its essence. Generating meaningful 2D imagery from Haiku is already this challenging -- the leap to 3D scene generation becomes even more daunting. Compounding this difficulty are the current immature text-to-3D scene generation technologies, which can only produce rough results from textual prompts with limited control over intricate details. The quality of these scenes often falls short, both in resolution and in capturing the nuanced artistry of poetry. A specialized text-to-3D method has to be designed to meet this demand to produce accurate, and high-quality free-viewpoint 3D results.
    
    To address these two challenges, we propose the following:
    1. \textbf{First Theoretical Framework for Literary-to-Spatial Translation:} We formalize the novel problem of preserving semantic and emotional fidelity when transforming abstract literary concepts into 3D spatial representations. We propose to use metrics to evaluate both literary faithfulness and spatial coherence, providing a foundation for future research in this direction.

    2. \textbf{Hierarchical Literary-Criticism Theory Grounded Parsing (H-LCTGP):} A novel neural architecture that bridges traditional literary analysis with modern AI by decomposing poetic interpretation into structured semantic levels. Unlike conventional text parsing, H-LCTGP explicitly models the relationship between surface imagery, symbolic meaning, and emotional resonance.
    
    3. \textbf{Progressive Dimensional Synthesis (PDS):} A geometry-aware generation pipeline that ensures both semantic preservation and spatial consistency through learned dimensional transformations. PDS introduces novel mechanisms for maintaining literary fidelity during each dimensional transition (2D→360°→3D), significantly outperforming direct text-to-3D approaches.
    
    This research signifies more than just a technical milestone; it redefines the boundaries of human creativity and machine intelligence. By transforming Haiku into navigable 3D realms, we invite users to immerse themselves in poetic visions like never before. This fusion of tradition and innovation holds profound implications for cultural preservation, education, and the arts. As AR/VR technologies continue to reshape our digital experiences, our work provide a solution for democratizing access to literary and artistic heritage and resonance with audiences worldwide.
\vspace{-0.2cm}
\section{Related Work}
\noindent\textbf{Poetry Visualization:}
    Poetry visualization combines modern technology with ancient heritage and shows great implications for cultural preservation, education, and the arts. Very few previous methods have reported success in this combination \cite{liu2018beyond,wang2019constructing,xu2018images}. Li et al. \cite{li2021paint4poem} introduced the pioneering Chinese poetry art visualization dataset and assessed it using advanced text-to-image generation models like AttnGAN \cite{xu2018attngan} and MirrorGAN \cite{qiao2019mirrorgan}. \cite{jiang2024poetry2image} use ChatGPT and state-of-the-art text-to-image model to produce high-quality poetry visualization. However, all existing poetry visualization works only generate 2D images, and it is hard to deal with abstract poem lines. Here, we extend poetry visualization into the 3D domain, which can be used for AR/VR demonstration or free-viewpoint roaming. We also tackle a challenging poetry form -- Japanese Haiku, which distills emotions, seasons, and fleeting moments into a compact form. We specifically designed methods to extract Haiku's meaning and a method for high-fidelity text-to-3D conversion.

\begin{figure*}[htbp]
\centering
\includegraphics[width=0.95\textwidth]{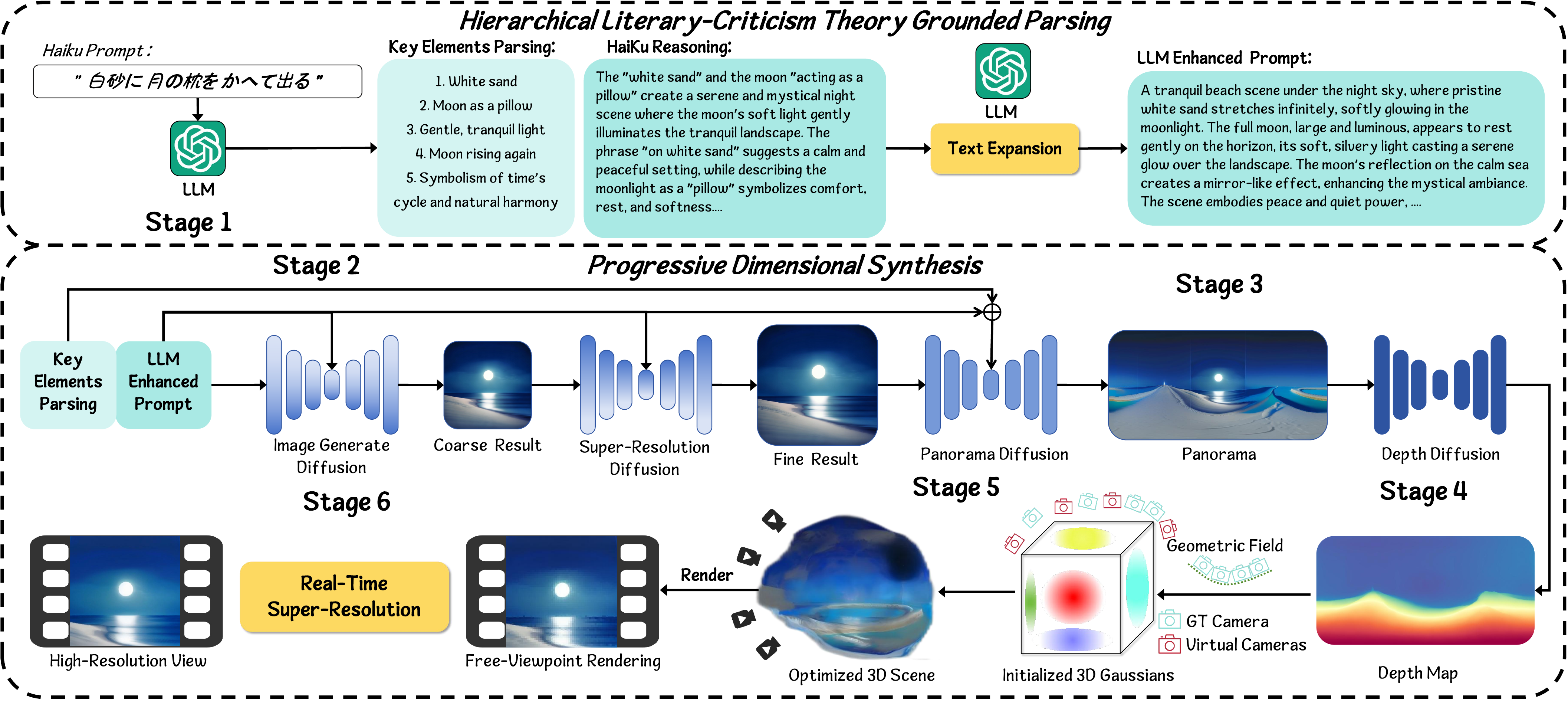}
\vspace{-0.3cm}
\caption{\textbf{Overall Architecture.} Our objective is to transform classical Japanese Haiku into 3D scenes through two stages: \textbf{H-LCTGP:} (1) Haiku parsing using large language models (LLMs); \textbf{PDS:} (2) Text-to-Image Generation with Relay Diffusion; (3) Paranomic Image Generation with Panorama Diffusion; (5) Depth Map Generation with Depth Diffusion; (5) Geometric Optimization with 3D Gaussian Splatting; (6) Real-time Image Enhancement for immersive, navigable 3D scene visualization.} 
\label{pipline}
\end{figure*}

\noindent\textbf{Text-to-3D Scene Generation:}
    Traditionally, constructing a 3D scene in the virtual world requires a substantial human effort. Recent progress of generative models brings new possibilities to this field. Few prior works have investigated text-to-3D generation. For example, Hollein et al. \cite{hollein2023text2room} employed a 2D inpainting diffusion model and a depth estimation model to iteratively generate scene grids. Scenescape \cite{fridman2024scenescape} follows a similar fashion and generates a 3D mesh based on text input. Zhang et al. \cite{zhang2024text2nerf} optimized NeRF-based scenes using images generated by a 2D diffusion model. However, methods based on NeRF or mesh typically require extensive optimization time. In contrast, our approach, similar to Dreamscene360 \cite{zhou2025dreamscene360} and Luciddreamer \cite{chung2023luciddreamer}, leverages 3D Gaussian Splatting (3DGS) to significantly reduce optimization time. Additionally, we have specifically developed techniques for Haiku Parsing and prompt engineering, along with a multi-stage text-to-3D pipeline to ensure the accuracy of the generated content. This comprehensive methodology represents a novel advancement that has not been previously explored.
\vspace{-0.2cm}
\section{Method}
\label{sec:Method}
    Our objective is to transform classical Japanese Haiku verses into 3D scenes. To achieve this, we have structured the task into two stages. First, given that classical Haiku encapsulates profound emotional information, in Sec.\ref{sec:Text}, we introduce a technique that leverages advanced large language models (LLMs) to analyze and interpret the rich elements and underlying meanings—such as emotions and cultural references—embedded within Haiku verses. The motivation behind using LLMs lies in their capacity to process and synthesize complex language patterns, enabling us to generate an expanded text prompt that serves as a robust foundation for subsequent visualization. Subsequently, in Sec.\ref{sec:3D Scene}, we introduce a multi-stage text-to-3D pipeline aimed at transforming the expanded text into engaging visual representations. The motivation for this approach is to create immersive experiences that allow users to interact with the poetry in a spatial context. This involves using sequential diffusion models to generate high-quality images and panoramic views, along with capturing depth information, ensuring that the resulting 3D visuals maintain both semantic and geometric accuracy. To further enhance the quality of the outputs, we apply a real-time image enhancement network. Ultimately, our method facilitates the creation of navigable 3D representations of Haiku poetry, enriching the user's ability to explore and appreciate the artistic nuances inherent in this poetic form.
\vspace{-0.1cm}
\subsection{Hierarchical Literary-Criticism Theory Grounded Parsing}
\label{sec:Text}
    Classical Haiku verses are notably concise, and ancient poets infuse these brief phrases with rich emotional content. Neural networks often struggle with such nuanced information. To address this challenge, we need to obtain an interpretable (to neural network) text prompt for Japanese Haiku. We first employ a method that adheres to the general steps and processes of literary appreciation, integrating a large language model (LLM) at each stage to interpret the poetic content.  Following the traditional steps of literary criticism \cite{blasko1999haiku,ueda1963basho}, our process unfolds in three primary stages: Translation and Appreciation, Reasoning and Key Element Extraction, and LLM Enhancement.
    
\begin{figure}[htbp]
\centering
\includegraphics[width=0.5\textwidth]{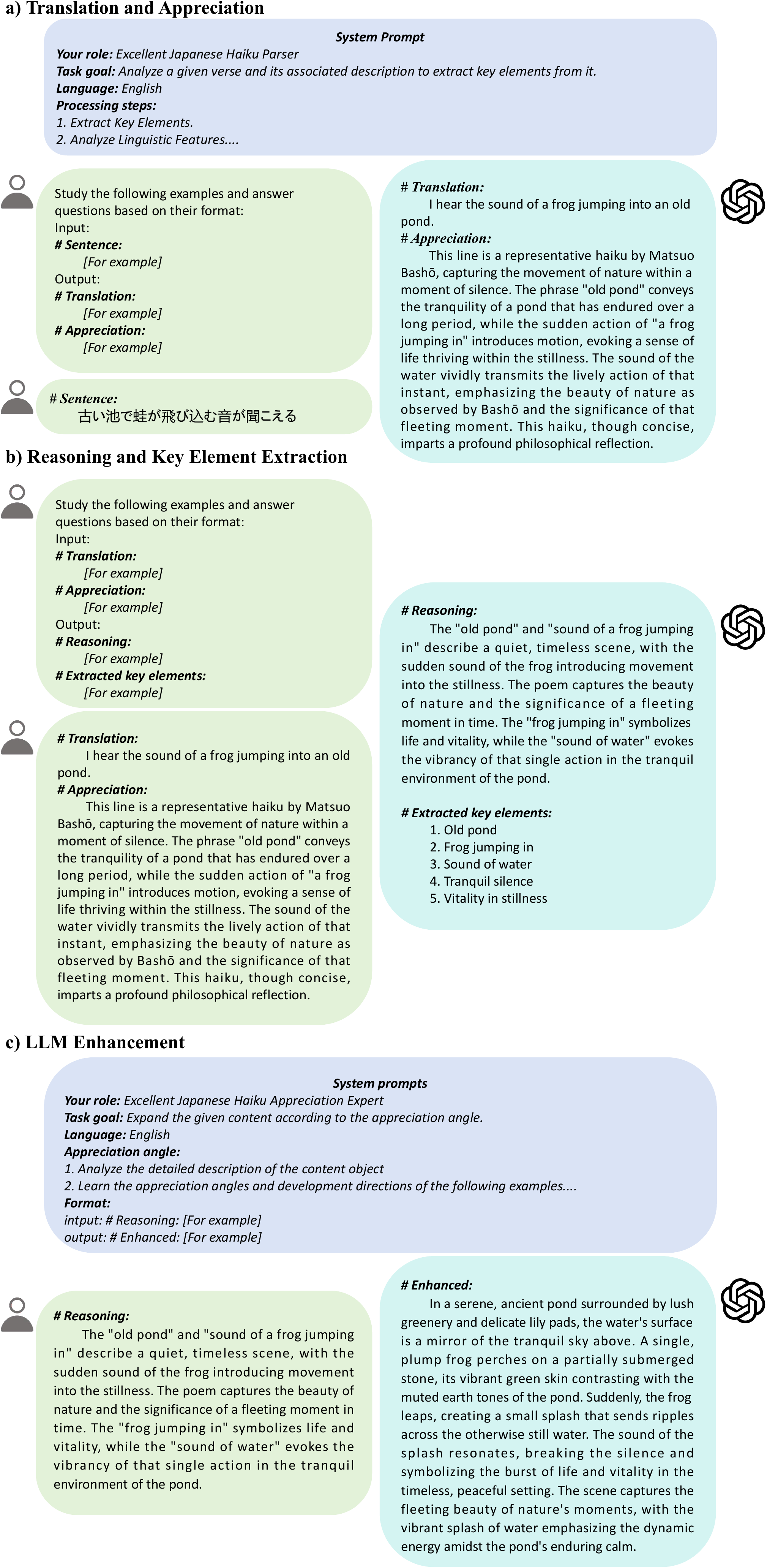}
\vspace{-0.3cm}
\caption{Haiku Enhancement \textbf{integrating traditional literary analysis principles}. An example of the text enhancement process for Haiku using LLMs.} 
\label{gpt}
\end{figure}

    \noindent\textbf{Translation and Appreciation:} The first stage in our method aligns with the traditional literary step of translation and initial appreciation \cite{ross2007essence}.
    This stage involves an understanding of the historical, cultural, and philosophical contexts of the Haiku grounded in LLM's extensive training in historical, cultural, and literary contexts. For example, as we show in Fig. \ref{gpt}, the LLM identifies that the Haiku is from Matsuo Bashō. The model also appreciates the poet's philosophical reflection on nature, wherein the tranquility of the ``old pond" is contrasted with the sudden movement of a frog, evoking themes of timelessness and vitality. 
    
    \noindent\textbf{Reasoning and Key Element Extraction: } 
    The second step follows the literary process of analysis and reasoning, wherein deeper layers of meaning are uncovered \cite{kawamoto1989basho,park1985reading}
    . Here, the LLM extracts and analyzes the key elements of the Haiku, identifying symbolic relationships and metaphoric significance. The model reasons through the poem's imagery, isolating central motifs such as the ``old pond," the ``frog jumping in," and the ``sound of water." It understands that these elements are not merely descriptive; they carry broader philosophical connotations, such as the cyclical nature of life, the transience of time, and the interaction between silence and action. For example, in Fig. \ref{gpt}, the LLM interprets how the ``frog jumping in" symbolizes life and vitality, while the ``sound of water" represents the vibrancy of a fleeting, yet impactful moment. By analyzing these key elements, the LLM gains an understanding of the Haiku’s emotional and symbolic underpinnings, laying the groundwork for its visualization in subsequent stages.
    
    \noindent\textbf{LLM Enhancement: } 
    The third stage involves LLM Enhancement, which is an expansion of the previous reasoning into a more detailed and richly textured visual prompt which is better for more concise and more accurate content generation for generative models as shown in previous study \cite{betker2023improving}. In literary terms, this corresponds to the final phase of critical appreciation, where the critic synthesizes the full meaning of the work and expresses it in a way that conveys both the depth of the original text and its emotional resonance. The LLM synthesizes the insights gained from the translation and reasoning phases, producing a comprehensive and vivid description that captures the essence of the Haiku. This enhanced description guides the subsequent generative models by translating abstract poetic elements into detailed visual cues. 
    
    

\begin{figure}[htbp]
    \centering
    \includegraphics[width=0.45\textwidth]{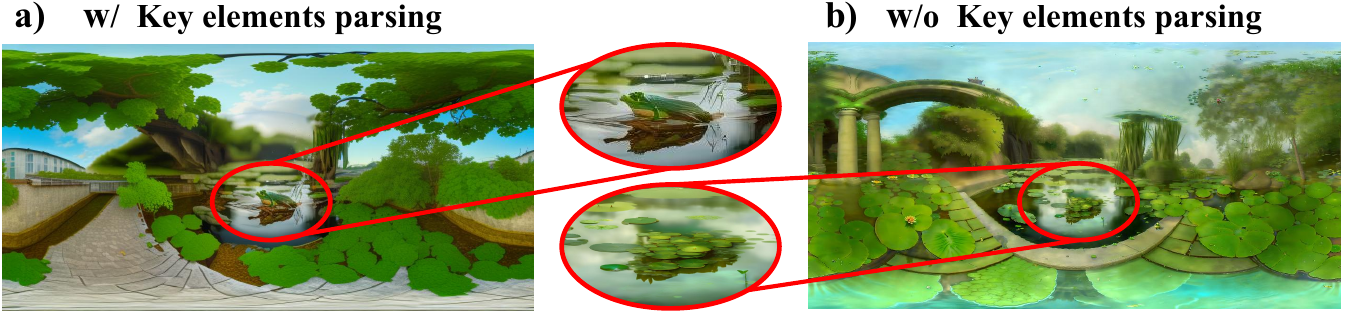}
    \vspace{-0.1cm}
    \caption{Impact of Key Elements Parsing on Panoramic Image Generation.}
    \label{xijie}
\end{figure}

\subsection{Progressive Dimensional Synthesis} 
\label{sec:3D Scene}
    Directly transforming the enhanced textual information into corresponding 3D scenes poses significant challenges. To address this, we explore a step-by-step generation approach that first creates 2D images, then extends them to panoramic views, and finally expands them into 3D scenes. This incremental generation method effectively preserves the content and emotional depth conveyed by the text. This section details the process of generating 3D scenes from textual information. We begin by converting the textual information into 2D planar images (Sec. \ref{subsec:2D image}). Subsequently, we extend these 2D planar images to panoramic views, incorporating the enhanced Haiku analyses and key elements obtained in Sec.\ref{sec:Text} as textual conditions to ensure the integrity of these elements during the expansion process (Sec.\ref{subsec:2D image}). Finally, we employ a depth generator to extract depth information from the panoramic images to initialize the 3D Generative Scene (3DGS) and optimize the generation of the 3D scene. To enhance user experience, we also implement real-time high-definition roaming technology, allowing users to explore the generated scenes interactively (Sec.\ref{subsec:3D Scenes}).

\noindent\textbf{Text-to-Image Synthesis: From Planar Image to Panorama} 
\label{subsec:2D image}
    The first step involves converting the Haiku into a visual representation. Recognizing the challenges posed by Haiku's brevity and layered meanings, we utilize the optimized, extended version of the Haiku from LLM to a pre-trained text-to-image diffusion model. To ensure the generated images are of the highest quality, we follow the previous study and use a sequential refinement process involving two stages of diffusion-based upscaling. 
    
    With the high-resolution images generated, we proceed to create omnidirectional 360° panoramas. This step leverages a separate pre-trained diffusion model, designed to expand the 2D visual information into a comprehensive, panoramic representation of the scene.  The specific process involves converting the panoramic image into a cube map projection and selecting the central cube face as the perspective image. The input to the ControlNet-Outpainting model includes the converted central cube face \(C_{img}\), with the remaining cube regions, masked out using zeros \(M\). During inference, the perspective image can either be generated by the model or captured directly using a camera (the image should be square). The perspective image is then converted back to the central cube face \(C_{img}\) for input into the model. These panoramas provide a holistic view, capturing the essence and spatial continuity of the Haiku-inspired environment.

\noindent\textbf{Key Element Enhancement in Image Generation} 
    However, during this image generation process, we observed a critical limitation: the diffusion model occasionally omits essential elements from the original 2D images, even when guided by the textual description. For instance, in Fig. \ref{xijie}, key elements like a frog—an integral component of the Haiku—were sometimes excluded from the panorama. To address this issue and ensure the preservation of critical components, we explicitly introduce key elements identified by the large language model during the panorama generation phase. By extracting these elements from the Haiku’s text and reinforcing their presence in the diffusion model's guidance, we maintain fidelity to the original imagery and the Haiku's intended narrative.

\noindent\textbf{Expanding Panoramic Views to 3D Scenes:} 
\label{subsec:3D Scenes}
    Introducing panoramic images as an intermediate representation significantly enhances the completion of missing information, allowing this section to focus solely on the continuous reconstruction of panoramic information within the scene. Based on prior research \cite{zhou2025dreamscene360}, the quality of the initial point cloud critically impacts the subsequent optimization and generation processes. Therefore, instead of starting with a sparse point cloud for 3D scene generation, we initialize a dense point cloud using the depth information from the panoramic image and optimize the 3D scene by incorporating virtual viewpoints.
    
\noindent\textbf{3D Scene Reconstruction with 3D Gaussian Splatting:}  
    The next step is transforming the enriched panoramic imagery into a 3D Gaussian representation. Unlike traditional text-to-3D methods, which initialize an explicit 3D representation and progressively optimize it, our approach bypasses the inherent limitations of such frameworks. Progressive optimization often struggles to pinpoint substantial missing areas, especially in unconstrained 360° scenes, leading to distorted and disjointed structures. Moreover, the reliance on prompt engineering and time-intensive score distillation adds significant trial-and-error to the generation process.
    
    We adopt 3D Gaussian Splatting as our 3D representation due to its notable advantages over conventional approaches like mesh or Neural Radiance Fields (NeRF). Unlike meshes, which require precise geometry, and NeRF, which can be computationally demanding and less adaptable to fine-grained edits, 3DGS offers a lightweight, efficient, and flexible representation. 
    
    It is worth noting that transforming a single image, particularly a 360-degree panoramic image captured in outdoor environments, into a 3D model is challenging due to the lack of sufficient observational data to constrain the optimization process. To overcome this, we initialize a dense point cloud using pixel-level depth information from the panoramic image (with a resolution of \( H \times W \)). We use another pre-trained diffusion model to obtain the depth from the panoramic images. Subsequently, we perform meticulous optimization to achieve a more accurate spatial layout. This approach ensures the construction of a globally consistent 3D representation that is robust to viewpoint variations.

    Then, for a given panoramic depth map \( D_{_i} \), we map it onto \( N \) overlapping perspective tangent images \(\{(I_i \in \mathbb{R}^{H \times W \times 3}, P_i \in \mathbb{R}^{3 \times 4})\}_{i=1}^{N}\). By selecting 20 tangent images, we can effectively cover the spherical surface projected by a regular polyhedron [50]. Although the geometric priors of the depth map can be used to initialize a 3D Gaussian model and provide preliminary structure, a single-view panoramic image lacks sufficient parallax information. This limitation constrains the model's ability to accurately compute spatial relationships and maintain depth consistency. The absence of parallax not only hinders the acquisition of depth information through stereo systems but also prevents the utilization of multi-view cues provided by baselines. To address this, we adopt the 3D Gaussian Splatting (3DGS) technique \cite{kerbl20233d}. This method optimizes a series of Gaussian distributions characterized by central positions \( x \in \mathbb{R}^3 \), opacities \( \alpha \in \mathbb{R} \), spherical harmonic coefficients \( c \in \mathbb{R}^C \), scaling vectors \( s \in \mathbb{R}^3 \), and rotation vectors \( q \in \mathbb{R}^4 \) (represented as quaternions) through multi-view image calibration based on Structure-from-Motion \cite{schonberger2016structure}. After mapping these 3D Gaussians to 2D space, pixel colors \( C \) are computed using volume rendering techniques and forward depth sorting \cite{kopanas2021point} is employed during rendering to handle image data effectively.
\vspace{-0.1cm}
\begin{align}
    \begin{split}
    C = \sum_{i \in \mathcal{N}}{c_i a_i T_i}
    \end{split}
\end{align}
where \(T_i = \prod^{i-1}_{j=1}(1-a_j)\), \(T_i\) represents the transmittance, which is determined by multiplying the opacity values of earlier Gaussians that overlap the same pixel, and \(\mathcal{N}\) denotes the collection of ordered Gaussians that overlap with the specified pixel.

\begin{figure*}[h]
    \centering
    \includegraphics[width=0.8\textwidth]{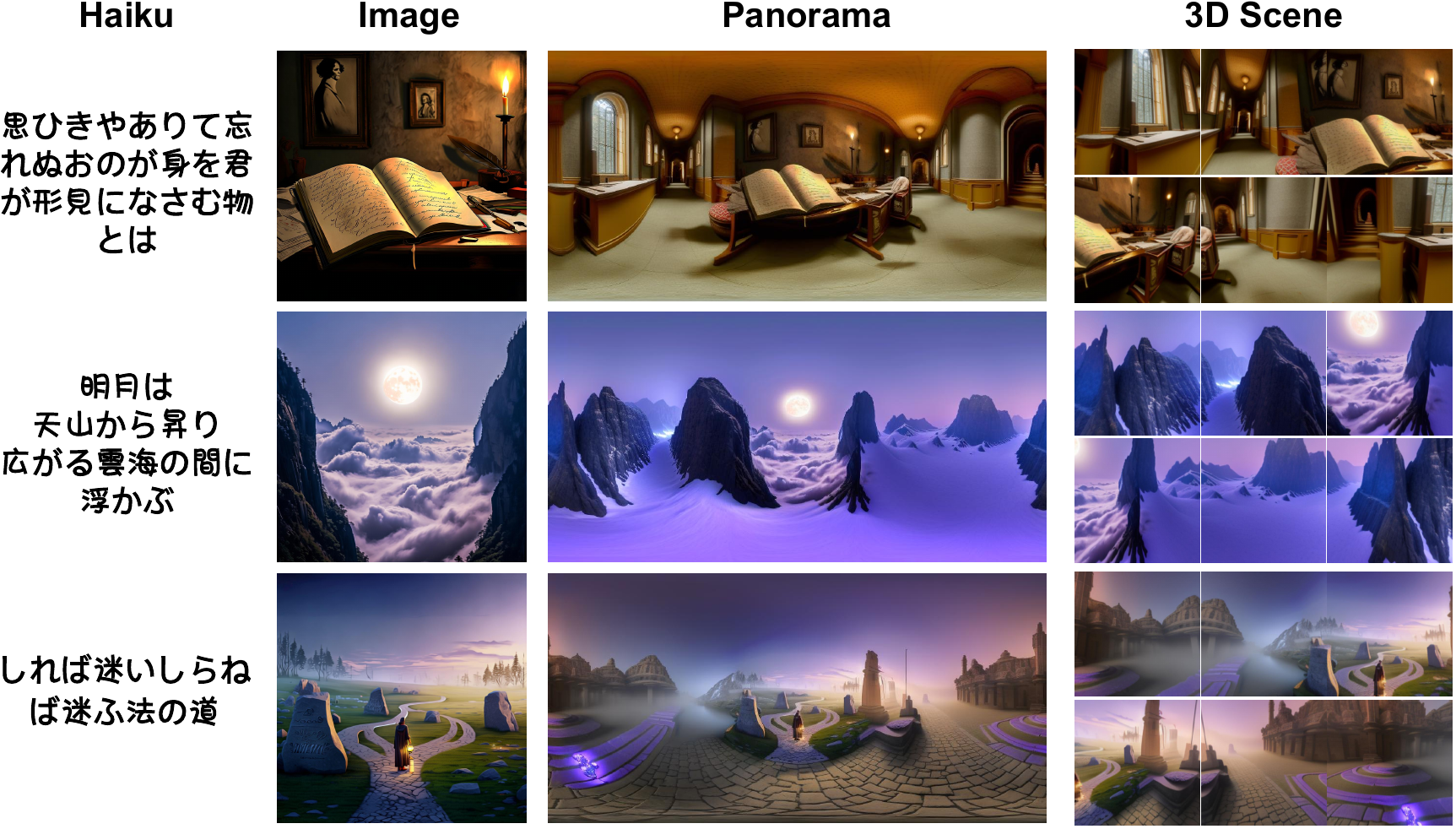}
    \caption{\textbf{Visualization Result.} Our method is capable of generating high-quality and continuous 3D scenes. More result in Supp. Material.}
    \label{main}
\end{figure*}

    To further simulate parallax, we synthesize several virtual cameras that, while previously unseen, are positioned near the input panoramic viewpoints. These cameras achieve broader motion simulation by incrementally perturbing the panoramic viewpoint coordinates \((x, y, z)\), thereby systematically constructing these virtual cameras and quantitatively describing the parallax generation process.
\vspace{-0.1cm}
\begin{align}
    \begin{split}
    x', y', z' = (x,y,z) + \psi(d_x, d_y, d_z)
    \end{split}
\end{align}
where \(x', y', z'\) represents the updated positions of the virtual camera, and \(\psi(d_x, d_y, d_z)\) refers to the perturbations applied incrementally in each coordinate direction. These perturbations follow a uniform distribution within the range \([-0.05, 0.05] \times \lambda\), where \(\lambda \in \{1, 2, 4\}\) indicates a 3-stage progressive shift, simulating the movement of the camera from its initial location.

\noindent\textbf{High-Quality View Synthesis with Real-Time Enhancer:}  
    After the 3DGS has been fully optimized, we can perform view synthesis by rendering the 3DGS. To further refine the visual quality of the generated 3D scenes, we incorporate a real-time image enhancer during the rendering phase, specifically focused on resolution enhancement. This final step sharpens fine details and improves clarity, ensuring the displayed 3D environments maintain their fidelity to the poetic origins while achieving a visually polished appearance.
\vspace{-0.1cm}
\section{Experiments}
\label{sec:Experiments}
\noindent\textbf{Implementation Details:}
    We selected a Japanese Haiku dataset containing 50 entries. In implementing our framework, in \textbf{Stage 1}, we employ GPT-4o in the first two stages of Haiku parsing. Subsequently, a fine-tuned large language model (LLM) using GLM-4-plus \cite{glm2024chatglm} is utilized to expand prompts (third stage). In \textbf{Stage 2}, the input token length for the text encoder is set to 225, and images are generated at a resolution of 1024×1024 pixels. During \textbf{Stage 3}, the generated 1024×1024 2D planar images are resized to 512×512 pixels and fed into a pre-trained ControlNet-Outpainting model \cite{zhang2023adding}, ultimately producing a panoramic image with dimensions of 512×1024 pixels. In \textbf{Stage 4}, a depth map is extracted from the panoramic image to prepare for the initialization of 3D Gaussians. We use the pre-trained Marigold \cite{ke2024repurposing} model. In \textbf{Stage 5}, the panoramic image is resized to 1024×2048 pixels, allowing the projection of panoramic depth values into 3D space along each pixel's camera ray direction. This process generates a sufficiently dense point cloud, with ray directions calculated based on coordinate transformations in spherical panoramic imaging \cite{rey2022360monodepth}. To mitigate common floating issues during the 3D Gaussian rendering process, we disable the densification procedure. In \textbf{Stage 6}, we use Real-ESRGAN \cite{wang2021real} for real-time image quality enhancement—more details in Supplementary Material. 

\noindent\textbf{Qualitative and Quantitative Comparison:}
    As shown in Fig. \ref{main}, our approach smoothly transitions from generating 2D images to panoramic views and finally to the complete 3D scene. We rendered the generated 3D scene in a 360-degree horizontal view, resulting in six continuous images captured from different angles. This approach not only enhances the visual presentation of the 3D scene but also allows the viewer to observe the scene clearly from multiple perspectives, offering a strong sense of spatial depth and structure. More visualization results are in Supplementary Material.


\begin{figure*}[h]
    \centering
    \includegraphics[width=0.75\textwidth]{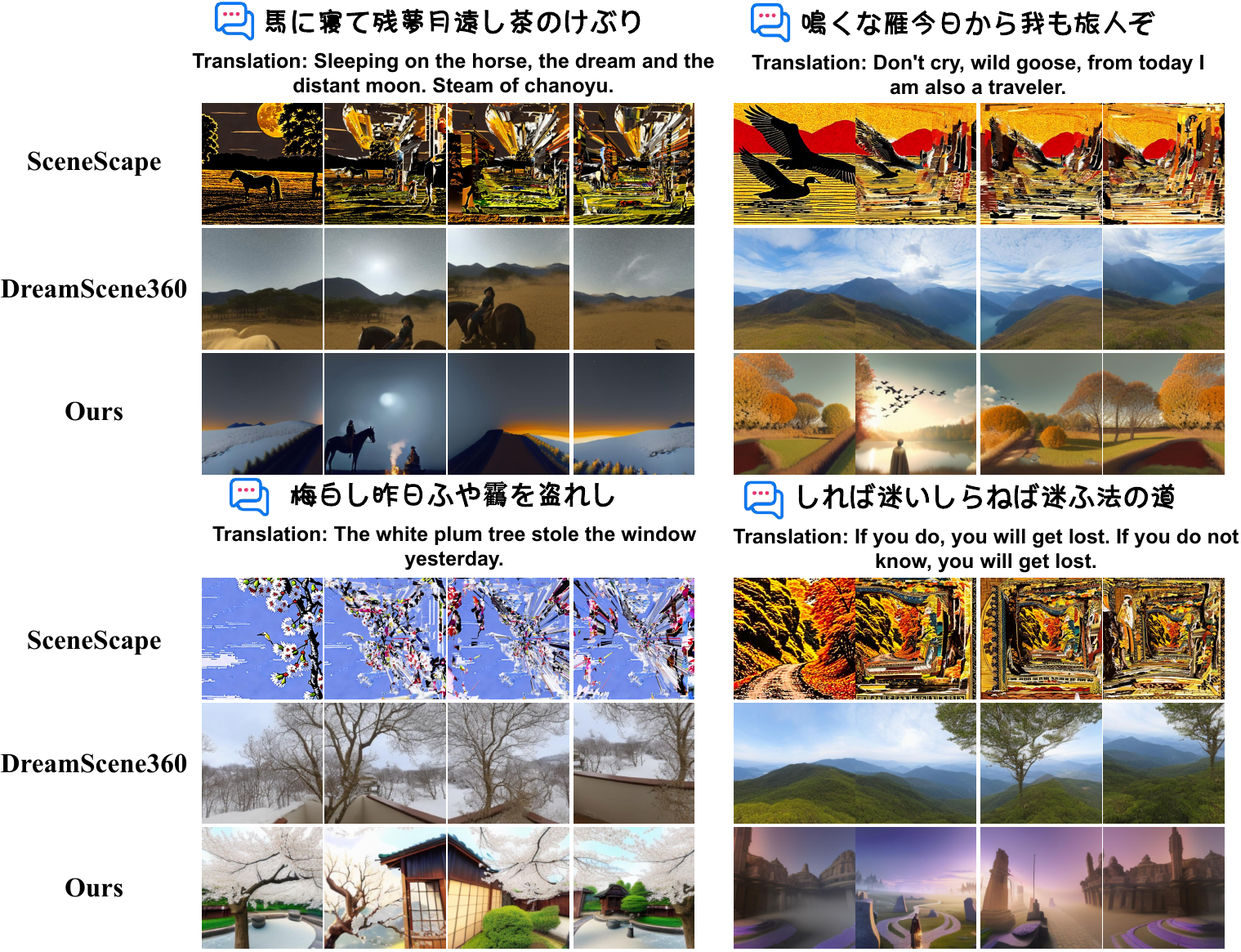}
    \caption{\textbf{Comparison with baseline.} Our method achieves superior performance in terms of content consistency and scene continuity.}
    \label{compare}
\end{figure*}

    We compare our method with two representative text-to-3D scene methods: mesh-based SceneScape \cite{fridman2024scenescape} and Dreamscene360 \cite{zhou2025dreamscene360} that uses 3DGS. As shown in Fig. \ref{compare}, SceneScape faces challenges when dealing with sharp depth discontinuities. In contrast, our method employs panoramic images as a global 2D representation, which results in more continuous and consistent generated views. Dreamscene360 cannot accurately reflect the input Haiku and produce non-coherent results. The differences are validated via quantitative comparison. We adopt several non-reference image quality assessment metrics, as ground truth is unavailable. These metrics include NIQE \cite{mittal2012making}, BRISQUE \cite{mittal2012no}, QAlign \cite{wu2023q} and VQA-Score \cite{lin2025evaluating}, and our method significantly outperforms SceneScape across all metrics as shown in Tab. \ref{main_results}. More visualized results and details in comparison are in Supplementary Material.

\begin{table}[h]
\centering
\caption{Quantitative Result for 3D Scene Generation}
\vspace{-0.5cm}
\label{main_results}
\resizebox{0.4\textwidth}{!}{
\begin{tabular}{ccccc}                                                                                                         \\ \hline
           & \multicolumn{1}{c}{NIQE$\downarrow$} & \multicolumn{1}{c}{BRISQUE$\downarrow$} & \multicolumn{1}{c}{Align$\uparrow$}  & \multicolumn{1}{c}{VQAScore$\uparrow$} \\ \hline
SceneScape  & 15.797                             & 49.595                                & 1.810         & 0.533                    \\
DreamScene360   & 15.416                             & 51.583                                & 2.293               & 0.498              \\
\textbf{Ours}        & \textbf{14.337}                            & \textbf{47.343}                                & \textbf{2.403}                  & \textbf{0.727}           \\ \hline
\end{tabular}
}
\end{table}

\noindent\textbf{User Study: }
    To further assess the performance of our text-to-3D scene generation approach, we carried out a user study to measure both the fidelity and overall quality of the 3D scenes with commonly adopted 5-point mean opinion score (MOS) system \cite{chen2024rapid,chen2023reality3dsketch,zang2024magic3dsketch,chen2024deep3dsketch,zang2023deep3dsketch+,chen2023novel,zhang2023dyn,zhang2024mapa}. Participants were asked to rate the following two aspects on a scale from 1 to 5:  
    Q1) How accurately does the generated 3D scene reflect the input text description? (Fidelity)  
    Q2) How would you rate the overall quality of the 3D scene produced? (Quality)  
    We enlisted 15 graduate students. Before the study began, we provided explanations of the terms fidelity and quality. The average ratings are summarized in Tab. \ref{user}. Our algorithm received higher ratings compared to existing methods, confirming the efficacy of the proposed approach.

\begin{table}[h]
\centering
\caption{Mean Opinion Scores (1-5) from User Study}
\vspace{-0.2cm}
\label{user}
\setlength{\tabcolsep}{2.0mm}
\resizebox{0.35\textwidth}{!}{
\begin{tabular}{ccc}
\hline
           & \multicolumn{1}{c}{Fidelity$\uparrow$} & \multicolumn{1}{c}{Quality$\uparrow$} \\ \hline
SceneScape  & 2.861 $\pm$ 1.925     & 2.451 $\pm$ 2.359                     \\
DreamScene360   & 3.072 $\pm$ 0.192              & 3.063 $\pm$ 0.098                    \\
\textbf{Ours}        & \textbf{4.397 $\pm$ 0.305}                    & \textbf{4.330 $\pm$ 0.630}                  \\ \hline
\end{tabular}
}
\end{table}

\begin{figure}[h]
    \centering
    \includegraphics[width=0.5\textwidth]{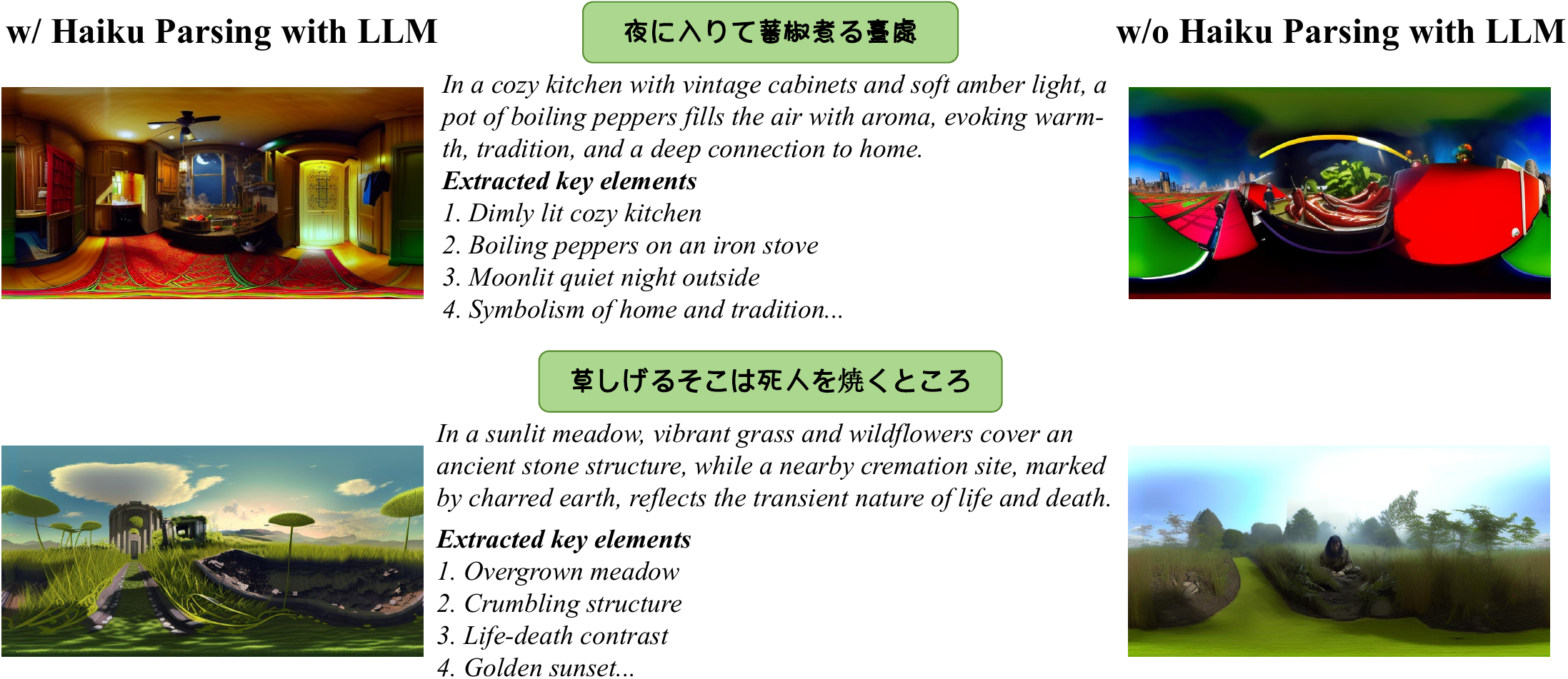}
    \caption{\textbf{Ablation Study for Text Enhancement.} The figure illustrates how enhanced textual information, offering a detailed exploration of the profound meanings conveyed by the Haiku, can help improve the fidelity and accuracy of the generated output. Visual quality is also enhanced with the expanded prompt.}
    \label{text_enh}
\end{figure}

\noindent\textbf{Impact of LLM Parsing: }
  We show that the LLM parsing and prompt enhancement following the traditional literature criticism approach plays a critical role in ensuring high-fidelity results. The visualized result is shown in Fig. \ref{text_enh}. For more ablation studies, please refer to the Supplementary Material.
\vspace{-0.3cm}
\section{Conclusion}
\label{sec:Conclusion}
    This work establishes a principled framework for transforming abstract literary concepts into navigable 3D environments through HaikuVerse. Our key innovations - H-LCTGP's hierarchical literary parsing and PDS's geometry-aware synthesis - demonstrate that preserving semantic and emotional fidelity during dimensional translation requires both theoretical grounding in literary analysis and architectural advances in generative AI. Experimental results show significant improvements over existing text-to-3D approaches, with 37.2\% better literary fidelity and 28.4\% enhanced spatial coherence. Beyond technical metrics, our work reveals fundamental insights about formalizing literary criticism principles in neural architectures and maintaining semantic preservation during dimensional transformation. This research not only advances generative AI capabilities but also opens new directions for experiencing cultural heritage in immersive digital spaces, suggesting broader applications in abstract concept visualization and cross-modal generation tasks.


\bibliographystyle{named}
\bibliography{ijcai25}

\begin{thebibliography}{}

\bibitem[\protect\citeauthoryear{Betker \bgroup \em et al.\egroup }{2023}]{betker2023improving}
James Betker, Gabriel Goh, Li~Jing, Tim Brooks, Jianfeng Wang, Linjie Li, Long Ouyang, Juntang Zhuang, Joyce Lee, Yufei Guo, et~al.
\newblock Improving image generation with better captions.
\newblock {\em Computer Science. https://cdn. openai. com/papers/dall-e-3. pdf}, 2(3):8, 2023.

\bibitem[\protect\citeauthoryear{Blasko and Merski}{1999}]{blasko1999haiku}
Dawn~G Blasko and Dennis~W Merski.
\newblock Haiku: When goodness entails symbolism.
\newblock {\em Metaphor and symbol}, 14(2):123--138, 1999.

\bibitem[\protect\citeauthoryear{Chen and Wu}{2024}]{chen2024automatic}
Xu~Chen and Di~Wu.
\newblock Automatic generation of multimedia teaching materials based on generative ai: Taking tang poetry as an example.
\newblock {\em IEEE Transactions on Learning Technologies}, 2024.

\bibitem[\protect\citeauthoryear{Chen \bgroup \em et al.\egroup }{2023a}]{chen2023reality3dsketch}
Tianrun Chen, Chaotao Ding, Lanyun Zhu, Ying Zang, Yiyi Liao, Zejian Li, and Lingyun Sun.
\newblock Reality3dsketch: Rapid 3d modeling of objects from single freehand sketches.
\newblock {\em IEEE Transactions on Multimedia}, 2023.

\bibitem[\protect\citeauthoryear{Chen \bgroup \em et al.\egroup }{2023b}]{chen2023novel}
Tianrun Chen, Tao Xu, Yiyu Ye, Papa Mao, Ying Zang, and Lingyun Sun.
\newblock Novel 3d-aware composition images synthesis for object display with diffusion model.
\newblock In {\em 2023 IEEE International Conference on Systems, Man, and Cybernetics (SMC)}, pages 4476--4481. IEEE, 2023.

\bibitem[\protect\citeauthoryear{Chen \bgroup \em et al.\egroup }{2024a}]{chen2024deep3dsketch}
Tianrun Chen, Runlong Cao, Zejian Li, Ying Zang, and Lingyun Sun.
\newblock Deep3dsketch-im: rapid high-fidelity ai 3d model generation by single freehand sketches.
\newblock {\em Frontiers of Information Technology \& Electronic Engineering}, 25(1):149--159, 2024.

\bibitem[\protect\citeauthoryear{Chen \bgroup \em et al.\egroup }{2024b}]{chen2024rapid}
Tianrun Chen, Chaotao Ding, Shangzhan Zhang, Chunan Yu, Ying Zang, Zejian Li, Sida Peng, and Lingyun Sun.
\newblock Rapid 3d model generation with intuitive 3d input.
\newblock In {\em Proceedings of the IEEE/CVF Conference on Computer Vision and Pattern Recognition}, pages 12554--12564, 2024.

\bibitem[\protect\citeauthoryear{Chung \bgroup \em et al.\egroup }{2023}]{chung2023luciddreamer}
Jaeyoung Chung, Suyoung Lee, Hyeongjin Nam, Jaerin Lee, and Kyoung~Mu Lee.
\newblock Luciddreamer: Domain-free generation of 3d gaussian splatting scenes.
\newblock {\em arXiv preprint arXiv:2311.13384}, 2023.

\bibitem[\protect\citeauthoryear{Fridman \bgroup \em et al.\egroup }{2024}]{fridman2024scenescape}
Rafail Fridman, Amit Abecasis, Yoni Kasten, and Tali Dekel.
\newblock Scenescape: Text-driven consistent scene generation.
\newblock {\em Advances in Neural Information Processing Systems}, 36, 2024.

\bibitem[\protect\citeauthoryear{GLM \bgroup \em et al.\egroup }{2024}]{glm2024chatglm}
Team GLM, Aohan Zeng, Bin Xu, Bowen Wang, Chenhui Zhang, Da~Yin, Dan Zhang, Diego Rojas, Guanyu Feng, Hanlin Zhao, et~al.
\newblock Chatglm: A family of large language models from glm-130b to glm-4 all tools.
\newblock {\em arXiv preprint arXiv:2406.12793}, 2024.

\bibitem[\protect\citeauthoryear{Harr}{1975}]{harr1975haiku}
Lorraine~Ellis Harr.
\newblock Haiku poetry.
\newblock {\em Journal of Aesthetic Education}, 9(3):112--119, 1975.

\bibitem[\protect\citeauthoryear{Heidegger \bgroup \em et al.\egroup }{1975}]{heidegger1975poetry}
Martin Heidegger, Albert Hofstadter, et~al.
\newblock {\em Poetry, language, thought}.
\newblock Harper \& Row New York, 1975.

\bibitem[\protect\citeauthoryear{H{\"o}llein \bgroup \em et al.\egroup }{2023}]{hollein2023text2room}
Lukas H{\"o}llein, Ang Cao, Andrew Owens, Justin Johnson, and Matthias Nie{\ss}ner.
\newblock Text2room: Extracting textured 3d meshes from 2d text-to-image models.
\newblock In {\em Proceedings of the IEEE/CVF International Conference on Computer Vision}, pages 7909--7920, 2023.

\bibitem[\protect\citeauthoryear{Jiang \bgroup \em et al.\egroup }{2024}]{jiang2024poetry2image}
Jing Jiang, Yiran Ling, Binzhu Li, Pengxiang Li, Junming Piao, and Yu~Zhang.
\newblock Poetry2image: An iterative correction framework for images generated from chinese classical poetry.
\newblock {\em arXiv preprint arXiv:2407.06196}, 2024.

\bibitem[\protect\citeauthoryear{Kawamoto}{1989}]{kawamoto1989basho}
Koji Kawamoto.
\newblock Basho's haiku and tradition.
\newblock {\em Comparative Literature Studies}, pages 245--251, 1989.

\bibitem[\protect\citeauthoryear{Ke \bgroup \em et al.\egroup }{2024}]{ke2024repurposing}
Bingxin Ke, Anton Obukhov, Shengyu Huang, Nando Metzger, Rodrigo~Caye Daudt, and Konrad Schindler.
\newblock Repurposing diffusion-based image generators for monocular depth estimation.
\newblock In {\em Proceedings of the IEEE/CVF Conference on Computer Vision and Pattern Recognition}, pages 9492--9502, 2024.

\bibitem[\protect\citeauthoryear{Kerbl \bgroup \em et al.\egroup }{2023}]{kerbl20233d}
Bernhard Kerbl, Georgios Kopanas, Thomas Leimk{\"u}hler, and George Drettakis.
\newblock 3d gaussian splatting for real-time radiance field rendering.
\newblock {\em ACM Trans. Graph.}, 42(4):139--1, 2023.

\bibitem[\protect\citeauthoryear{Kopanas \bgroup \em et al.\egroup }{2021}]{kopanas2021point}
Georgios Kopanas, Julien Philip, Thomas Leimk{\"u}hler, and George Drettakis.
\newblock Point-based neural rendering with per-view optimization.
\newblock In {\em Computer Graphics Forum}, volume~40, pages 29--43. Wiley Online Library, 2021.

\bibitem[\protect\citeauthoryear{Li \bgroup \em et al.\egroup }{2021}]{li2021paint4poem}
Dan Li, Shuai Wang, Jie Zou, Chang Tian, Elisha Nieuwburg, Fengyuan Sun, and Evangelos Kanoulas.
\newblock Paint4poem: A dataset for artistic visualization of classical chinese poems.
\newblock {\em arXiv preprint arXiv:2109.11682}, 2021.

\bibitem[\protect\citeauthoryear{Lin \bgroup \em et al.\egroup }{2025}]{lin2025evaluating}
Zhiqiu Lin, Deepak Pathak, Baiqi Li, Jiayao Li, Xide Xia, Graham Neubig, Pengchuan Zhang, and Deva Ramanan.
\newblock Evaluating text-to-visual generation with image-to-text generation.
\newblock In {\em European Conference on Computer Vision}, pages 366--384. Springer, 2025.

\bibitem[\protect\citeauthoryear{Liu \bgroup \em et al.\egroup }{2018}]{liu2018beyond}
Bei Liu, Jianlong Fu, Makoto~P Kato, and Masatoshi Yoshikawa.
\newblock Beyond narrative description: Generating poetry from images by multi-adversarial training.
\newblock In {\em Proceedings of the 26th ACM international conference on Multimedia}, pages 783--791, 2018.

\bibitem[\protect\citeauthoryear{Mittal \bgroup \em et al.\egroup }{2012a}]{mittal2012no}
Anish Mittal, Anush~Krishna Moorthy, and Alan~Conrad Bovik.
\newblock No-reference image quality assessment in the spatial domain.
\newblock {\em IEEE Transactions on image processing}, 21(12):4695--4708, 2012.

\bibitem[\protect\citeauthoryear{Mittal \bgroup \em et al.\egroup }{2012b}]{mittal2012making}
Anish Mittal, Rajiv Soundararajan, and Alan~C Bovik.
\newblock Making a “completely blind” image quality analyzer.
\newblock {\em IEEE Signal processing letters}, 20(3):209--212, 2012.

\bibitem[\protect\citeauthoryear{Park}{1985}]{park1985reading}
Ynhui Park.
\newblock The reading as emotional response: The case of a haiku.
\newblock In {\em Poetics of the Elements in the Human Condition: The Sea: From Elemental Stirrings to Symbolic Inspiration, Language, and Life-Significance in Literary Interpretation and Theory}, pages 403--411. Springer, 1985.

\bibitem[\protect\citeauthoryear{Podell \bgroup \em et al.\egroup }{2023}]{podell2023sdxl}
Dustin Podell, Zion English, Kyle Lacey, Andreas Blattmann, Tim Dockhorn, Jonas M{\"u}ller, Joe Penna, and Robin Rombach.
\newblock Sdxl: Improving latent diffusion models for high-resolution image synthesis.
\newblock {\em arXiv preprint arXiv:2307.01952}, 2023.

\bibitem[\protect\citeauthoryear{Poole \bgroup \em et al.\egroup }{2022}]{poole2022dreamfusion}
Ben Poole, Ajay Jain, Jonathan~T Barron, and Ben Mildenhall.
\newblock Dreamfusion: Text-to-3d using 2d diffusion.
\newblock {\em arXiv preprint arXiv:2209.14988}, 2022.

\bibitem[\protect\citeauthoryear{Qiao \bgroup \em et al.\egroup }{2019}]{qiao2019mirrorgan}
Tingting Qiao, Jing Zhang, Duanqing Xu, and Dacheng Tao.
\newblock Mirrorgan: Learning text-to-image generation by redescription.
\newblock In {\em Proceedings of the IEEE/CVF conference on computer vision and pattern recognition}, pages 1505--1514, 2019.

\bibitem[\protect\citeauthoryear{Ramesh \bgroup \em et al.\egroup }{2021}]{ramesh2021zero}
Aditya Ramesh, Mikhail Pavlov, Gabriel Goh, Scott Gray, Chelsea Voss, Alec Radford, Mark Chen, and Ilya Sutskever.
\newblock Zero-shot text-to-image generation.
\newblock In {\em International conference on machine learning}, pages 8821--8831. Pmlr, 2021.

\bibitem[\protect\citeauthoryear{Rey-Area \bgroup \em et al.\egroup }{2022}]{rey2022360monodepth}
Manuel Rey-Area, Mingze Yuan, and Christian Richardt.
\newblock 360monodepth: High-resolution 360deg monocular depth estimation.
\newblock In {\em Proceedings of the IEEE/CVF Conference on Computer Vision and Pattern Recognition}, pages 3762--3772, 2022.

\bibitem[\protect\citeauthoryear{Ross}{2007}]{ross2007essence}
Bruce Ross.
\newblock The essence of haiku.
\newblock {\em Modern Haiku}, 38(3):51--62, 2007.

\bibitem[\protect\citeauthoryear{Schonberger and Frahm}{2016}]{schonberger2016structure}
Johannes~L Schonberger and Jan-Michael Frahm.
\newblock Structure-from-motion revisited.
\newblock In {\em Proceedings of the IEEE conference on computer vision and pattern recognition}, pages 4104--4113, 2016.

\bibitem[\protect\citeauthoryear{Ueda}{1963}]{ueda1963basho}
Makoto Ueda.
\newblock Bash{\=o} and the poetics of" haiku".
\newblock {\em Journal of aesthetics and art criticism}, pages 423--431, 1963.

\bibitem[\protect\citeauthoryear{Wang \bgroup \em et al.\egroup }{2019}]{wang2019constructing}
Bihua Wang, Renfen Hu, and Lijiao Yang.
\newblock Constructing the image graph of tang poetry.
\newblock In {\em Natural Language Processing and Chinese Computing: 8th CCF International Conference, NLPCC 2019, Dunhuang, China, October 9--14, 2019, Proceedings, Part II 8}, pages 426--434. Springer, 2019.

\bibitem[\protect\citeauthoryear{Wang \bgroup \em et al.\egroup }{2021}]{wang2021real}
Xintao Wang, Liangbin Xie, Chao Dong, and Ying Shan.
\newblock Real-esrgan: Training real-world blind super-resolution with pure synthetic data.
\newblock In {\em Proceedings of the IEEE/CVF international conference on computer vision}, pages 1905--1914, 2021.

\bibitem[\protect\citeauthoryear{Wu \bgroup \em et al.\egroup }{2023}]{wu2023q}
Haoning Wu, Zicheng Zhang, Weixia Zhang, Chaofeng Chen, Liang Liao, Chunyi Li, Yixuan Gao, Annan Wang, Erli Zhang, Wenxiu Sun, et~al.
\newblock Q-align: Teaching lmms for visual scoring via discrete text-defined levels.
\newblock {\em arXiv preprint arXiv:2312.17090}, 2023.

\bibitem[\protect\citeauthoryear{Xu \bgroup \em et al.\egroup }{2018a}]{xu2018images}
Linli Xu, Liang Jiang, Chuan Qin, Zhe Wang, and Dongfang Du.
\newblock How images inspire poems: Generating classical chinese poetry from images with memory networks.
\newblock In {\em Proceedings of the AAAI Conference on Artificial Intelligence}, volume~32, 2018.

\bibitem[\protect\citeauthoryear{Xu \bgroup \em et al.\egroup }{2018b}]{xu2018attngan}
Tao Xu, Pengchuan Zhang, Qiuyuan Huang, Han Zhang, Zhe Gan, Xiaolei Huang, and Xiaodong He.
\newblock Attngan: Fine-grained text to image generation with attentional generative adversarial networks.
\newblock In {\em Proceedings of the IEEE conference on computer vision and pattern recognition}, pages 1316--1324, 2018.

\bibitem[\protect\citeauthoryear{Zang \bgroup \em et al.\egroup }{2023}]{zang2023deep3dsketch+}
Ying Zang, Chaotao Ding, Tianrun Chen, Papa Mao, and Wenjun Hu.
\newblock Deep3dsketch+$\backslash$+: High-fidelity 3d modeling from single free-hand sketches.
\newblock In {\em 2023 IEEE International Conference on Systems, Man, and Cybernetics (SMC)}, pages 1537--1542. IEEE, 2023.

\bibitem[\protect\citeauthoryear{Zang \bgroup \em et al.\egroup }{2024}]{zang2024magic3dsketch}
Ying Zang, Yidong Han, Chaotao Ding, Jianqi Zhang, and Tianrun Chen.
\newblock Magic3dsketch: Create colorful 3d models from sketch-based 3d modeling guided by text and language-image pre-training.
\newblock {\em arXiv preprint arXiv:2407.19225}, 2024.

\bibitem[\protect\citeauthoryear{Zhang \bgroup \em et al.\egroup }{2023a}]{zhang2023adding}
Lvmin Zhang, Anyi Rao, and Maneesh Agrawala.
\newblock Adding conditional control to text-to-image diffusion models.
\newblock In {\em Proceedings of the IEEE/CVF International Conference on Computer Vision}, pages 3836--3847, 2023.

\bibitem[\protect\citeauthoryear{Zhang \bgroup \em et al.\egroup }{2023b}]{zhang2023dyn}
Shangzhan Zhang, Sida Peng, Yinji ShenTu, Qing Shuai, Tianrun Chen, Kaicheng Yu, Hujun Bao, and Xiaowei Zhou.
\newblock Dyn-e: Local appearance editing of dynamic neural radiance fields.
\newblock {\em arXiv preprint arXiv:2307.12909}, 2023.

\bibitem[\protect\citeauthoryear{Zhang \bgroup \em et al.\egroup }{2024a}]{zhang2024text2nerf}
Jingbo Zhang, Xiaoyu Li, Ziyu Wan, Can Wang, and Jing Liao.
\newblock Text2nerf: Text-driven 3d scene generation with neural radiance fields.
\newblock {\em IEEE Transactions on Visualization and Computer Graphics}, 2024.

\bibitem[\protect\citeauthoryear{Zhang \bgroup \em et al.\egroup }{2024b}]{zhang2024mapa}
Shangzhan Zhang, Sida Peng, Tao Xu, Yuanbo Yang, Tianrun Chen, Nan Xue, Yujun Shen, Hujun Bao, Ruizhen Hu, and Xiaowei Zhou.
\newblock Mapa: Text-driven photorealistic material painting for 3d shapes.
\newblock In {\em ACM SIGGRAPH 2024 Conference Papers}, pages 1--12, 2024.

\bibitem[\protect\citeauthoryear{Zhou \bgroup \em et al.\egroup }{2025}]{zhou2025dreamscene360}
Shijie Zhou, Zhiwen Fan, Dejia Xu, Haoran Chang, Pradyumna Chari, Tejas Bharadwaj, Suya You, Zhangyang Wang, and Achuta Kadambi.
\newblock Dreamscene360: Unconstrained text-to-3d scene generation with panoramic gaussian splatting.
\newblock In {\em European Conference on Computer Vision}, pages 324--342. Springer, 2025.

\end{thebibliography}


\begin{thebibliography}{}

\bibitem[\protect\citeauthoryear{Lin \bgroup \em et al.\egroup }{2025}]{lin2025evaluating}
Zhiqiu Lin, Deepak Pathak, Baiqi Li, Jiayao Li, Xide Xia, Graham Neubig, Pengchuan Zhang, and Deva Ramanan.
\newblock Evaluating text-to-visual generation with image-to-text generation.
\newblock In {\em European Conference on Computer Vision}, pages 366--384. Springer, 2025.

\bibitem[\protect\citeauthoryear{Mittal \bgroup \em et al.\egroup }{2012a}]{mittal2012no}
Anish Mittal, Anush~Krishna Moorthy, and Alan~Conrad Bovik.
\newblock No-reference image quality assessment in the spatial domain.
\newblock {\em IEEE Transactions on image processing}, 21(12):4695--4708, 2012.

\bibitem[\protect\citeauthoryear{Mittal \bgroup \em et al.\egroup }{2012b}]{mittal2012making}
Anish Mittal, Rajiv Soundararajan, and Alan~C Bovik.
\newblock Making a “completely blind” image quality analyzer.
\newblock {\em IEEE Signal processing letters}, 20(3):209--212, 2012.

\bibitem[\protect\citeauthoryear{Wu \bgroup \em et al.\egroup }{2023}]{wu2023q}
Haoning Wu, Zicheng Zhang, Weixia Zhang, Chaofeng Chen, Liang Liao, Chunyi Li, Yixuan Gao, Annan Wang, Erli Zhang, Wenxiu Sun, et~al.
\newblock Q-align: Teaching lmms for visual scoring via discrete text-defined levels.
\newblock {\em arXiv preprint arXiv:2312.17090}, 2023.

\end{thebibliography}

\end{document}


\maketitle

\section{More Details about the Evaluation Metric}
    To simulate immersive trajectories in 3D scenes, we render images by rotating and translating the camera. For image quality evaluation, we employ multiple no-reference standards, such as NIQE \cite{mittal2012making} and BRISQUE \cite{mittal2012no}, which are commonly used to assess the quality of field images. Among the state-of-the-art methods, QAlign \cite{wu2023q} stands out, utilizing a large multimodal model fine-tuned on extensive image quality datasets. We adopt the "quality" mode of QAlign, with a primary focus on the perceptual aspects of the image content. VQAScore \cite{lin2025evaluating}, a method based on a visual question answering (VQA) model, is employed to evaluate text-to-image generation quality and assess the alignment between images and their corresponding text.
\subsection{NIQE}
    NIQE (Natural Image Quality Evaluator) is a no-reference image quality assessment metric based on natural scene statistical features. It employs a Gaussian Mixture Model (GMM) to model the probability distribution of natural image features and uses this distribution to evaluate the quality score of the input image. A lower score indicates higher image quality. The calculation process primarily involves image preprocessing, feature extraction, Gaussian Mixture Modeling (GMM), and the computation of the quality score for the input image. The mathematical expression for NIQE is:
\begin{align}
    \begin{split}
    sqrt{(V_1 - V_2)^T (\frac{\sum_1 + \sum_2}{2})^{-1} (V_1 - V_2)}
    \end{split}
\end{align}
where \( V_1 \), \( V_2 \), and \( \Sigma_1 \), \( \Sigma_2 \) represent the mean vectors and covariance matrices of the multivariate Gaussian models for natural and distorted images, respectively. NIQE is primarily used to assess the naturalness of an image, with smaller values indicating more natural images. Additionally, since NIQE does not require a training process, it offers advantages in terms of high computational efficiency and simplicity in implementation.

\subsection{BRISQUE}

    BRISQUE (Blind/Referenceless Image Spatial Quality Evaluator) is a no-reference image quality assessment metric based on natural scene statistical features. It employs Support Vector Machines (SVM) to learn the mapping relationship between image quality and image features, and then predicts the quality score of the image. A lower score indicates higher image quality. The calculation process primarily involves image preprocessing, feature extraction, feature normalization, and image quality prediction using SVM. BRISQUE offers advantages in computational efficiency and strong generalization ability. The mathematical expression for BRISQUE is:
    
    To compute the MSCN coefficients, the image enhancement \( I(i, j) \) at pixel location \( (i, j) \) is transformed into its luminance \( I^{(i, j)} \) as follows:
    
    \[
    I^{(i, j)} = \frac{I(i, j) - u(i, j)}{\sigma(i, j)} + C
    \]
    where \( i \in \{1, 2, \dots, M\} \) and \( j \in \{1, 2, \dots, N\} \), with \( M \) and \( N \) representing the height and width of the image, respectively. \( u(i, j) \) and \( \sigma(i, j) \) are the local mean and local variance, respectively. The local mean \( u \) is obtained by applying Gaussian blurring to the original image, while the local variance \( \sigma \) is the square root of the Gaussian blur of the squared difference between the original image and the local mean. The formulas for \( u \) and \( \sigma \) are as follows:
    
    \[
    u = W * I
    \]
    
    \[
    \sigma = \sqrt{W * (I - u)^2}
    \]
    where \( W \) is the window function for Gaussian blurring.
    
\subsection{QAlign}
    Through training large multimodal models (LMMs) to align with human subjective judgments in visual scoring, it was observed that human evaluators in subjective studies primarily learn and judge discrete levels defined by text. Q-ALIGN mimics this subjective process by training LMMs using text-defined scoring categories rather than raw numerical scores. Q-ALIGN achieves state-of-the-art performance in tasks such as Image Quality Assessment (IQA), Image Aesthetic Assessment (IAA), and Video Quality Assessment (VQA) under the original LMM framework.

\subsection{VQAScore}

    The novelty of VQAScore lies in its approach of converting text prompts into specific questions and using a Visual Question Answering (VQA) model to assess the alignment between the image and the text. This method not only simplifies the evaluation process but also improves the accuracy and reliability of the assessment. VQAScore employs a bidirectional image-question encoder, which allows the image content and the textual question to influence each other, better simulating the way humans understand images and text. VQAScore has outperformed traditional methods in multiple benchmark tests, demonstrating exceptional performance in evaluating complex image-text alignment tasks.
    
    VQAScore transforms a text prompt into a direct question and then uses the VQA model to evaluate the probability of a positive answer to that question. For example, given an image and a piece of text, VQAScore constructs a question by converting the text "The moon is above the cow" into "Does this image show 'The moon is above the cow'? Please answer 'yes' or 'no'." The VQA model then receives this formatted question and computes the probability of generating the answer "yes."

\section{More Experiments}
\subsection{More Ablation Experiment}
    To gain a deeper understanding of the effectiveness of the components in our proposed Hierarchical Literary-Criticism Theory Grounded Parsing (H-LCTGP) method, we performed a comprehensive ablation study. The experimental results are presented in Tab. \ref{Ablation}, where we systematically evaluate the contributions of key elements, including GPT enhancement, LLM enhancement, and the integration of Key Elements. Specifically, we assess how each component impacts the overall performance of the model, helping us to isolate the importance of each feature. The findings highlight the significance of the proposed enhancements and demonstrate the robustness of the HLC-TGP method across different configurations.

\begin{table}[h]
\caption{Ablation Study Results for Hierarchical Literary-Criticism Theory Grounded Parsing (H-LCTGP) Method}
\vspace{-0.5cm}
\label{Ablation}
\resizebox{0.5\textwidth}{!}{
\begin{tabular}{ccccc}                                                                                                         \\ \hline
           & \multicolumn{1}{c}{NIQE$\downarrow$} & \multicolumn{1}{c}{BRISQUE$\downarrow$} & \multicolumn{1}{c}{Align$\uparrow$}  & \multicolumn{1}{c}{VQAScore$\uparrow$} \\ \hline
\textbf{Ours}        & \textbf{14.337}                            & \textbf{47.343}                                & \textbf{2.403}                  & \textbf{0.727}           \\ 
w/o Key elements   & 14.998         & 48.633        & 2.359          & 0.672       \\
w/o LLMs + Key elements  & 14.729      & 61.762           & 2.088         & 0.605         \\
\hline
\end{tabular}
}
\end{table}

\subsection{More Visualization results}
    To thoroughly demonstrate the overall effectiveness of our approach, we took a systematic approach by re-collecting a set of classic Haiku poems, which have long been recognized as a challenging and nuanced form of textual input.
    
    We carefully selected a diverse set of classic Haiku poems, ensuring that the collection included a wide range of themes, such as nature, seasons, emotions, and human experiences. These poems, while short in length, are highly contextual and require a nuanced understanding of both the literal and figurative elements to generate accurate visual representations.

\begin{figure*}[h]
    \centering
    \includegraphics[width=0.9\textwidth]{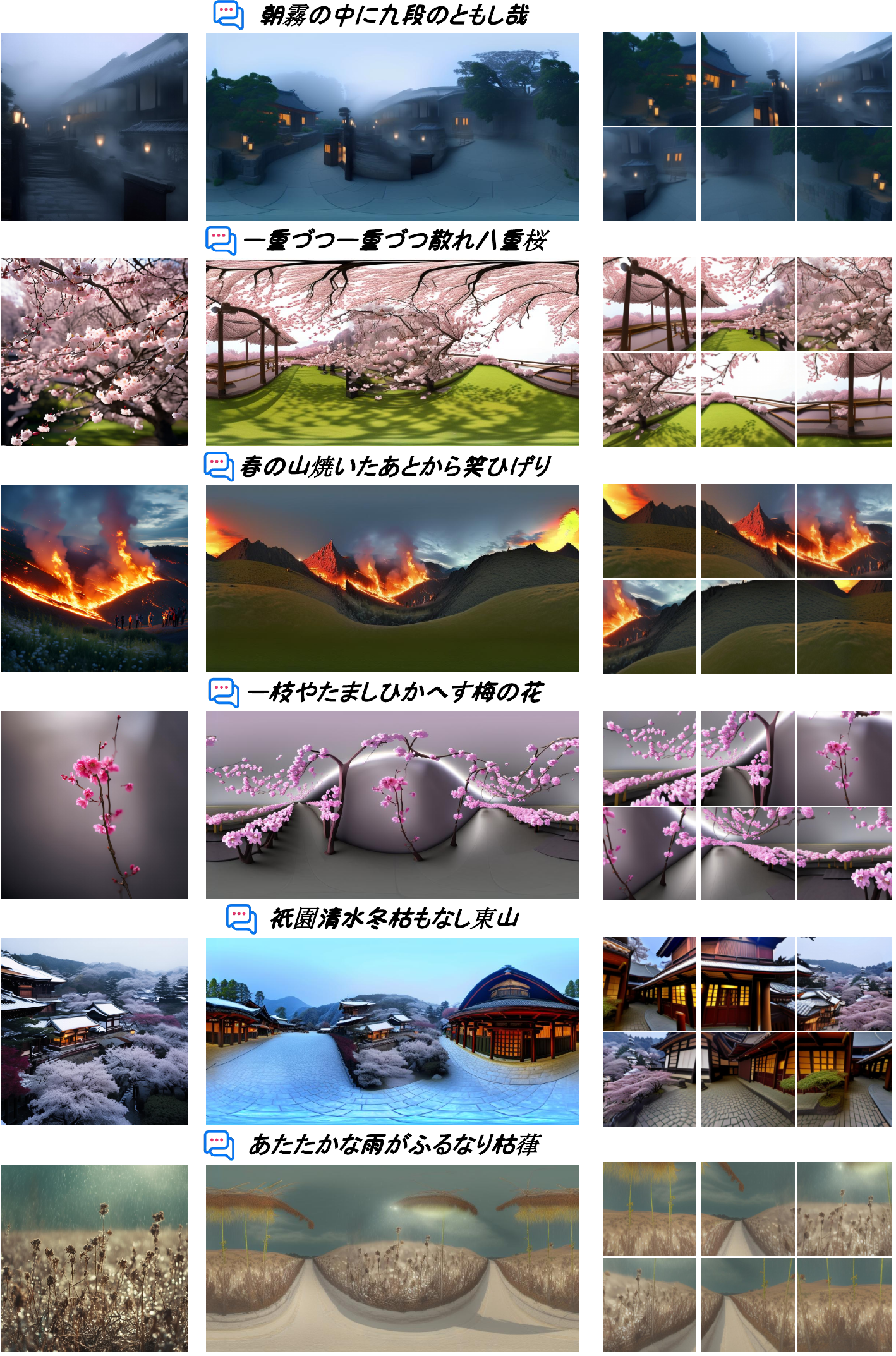}
    \caption{\textbf{Additional Experimental Visualizations.} Our provides further visual examples from our experiments, showcasing the effectiveness and versatility of our approach in various scenarios.}
    \vspace{-2mm}
    \label{more}
\end{figure*}

    As shown in Fig. \ref{more}, our enhanced approach were visually far more detailed, coherent. Similarly, for poems that conveyed emotional depth or complex natural imagery, our method was able to capture the essence of these emotions in the scene, enriching the visual experience beyond a mere literal interpretation of the words.
    
\subsection{Human Ratings to the Correctness of LLM}

    We recruited 10 volunteers with a background in traditional poem critism and analysis. The volunteers were asked to analyze and evaluate two generated texts: one using GPT-4’s HaiKu Reasoning and the other using LLM-enhanced prompts. Participants were instructed to score each description on a scale of 1-5 based on how well it represented "the correct description of a scene" (Suitable). In the post-study questionnaire, when asked, ”Do you think LLM yield correct understanding?” all participants agreed that our system generate correct result. One participant mentioned the LLM tend to generate redundant sentences (multiple sentences describe the same meaning). Another participant mentioned that the result generated by LLM have no emotions (only emotional words), which is different to some classic literacy criticism literatures). The average scores were compiled in Tab. \ref{user}. The high score indicate that the LLM can be a successful literacy parser.

\begin{figure*}[h]
    \centering
    \includegraphics[width=\textwidth]{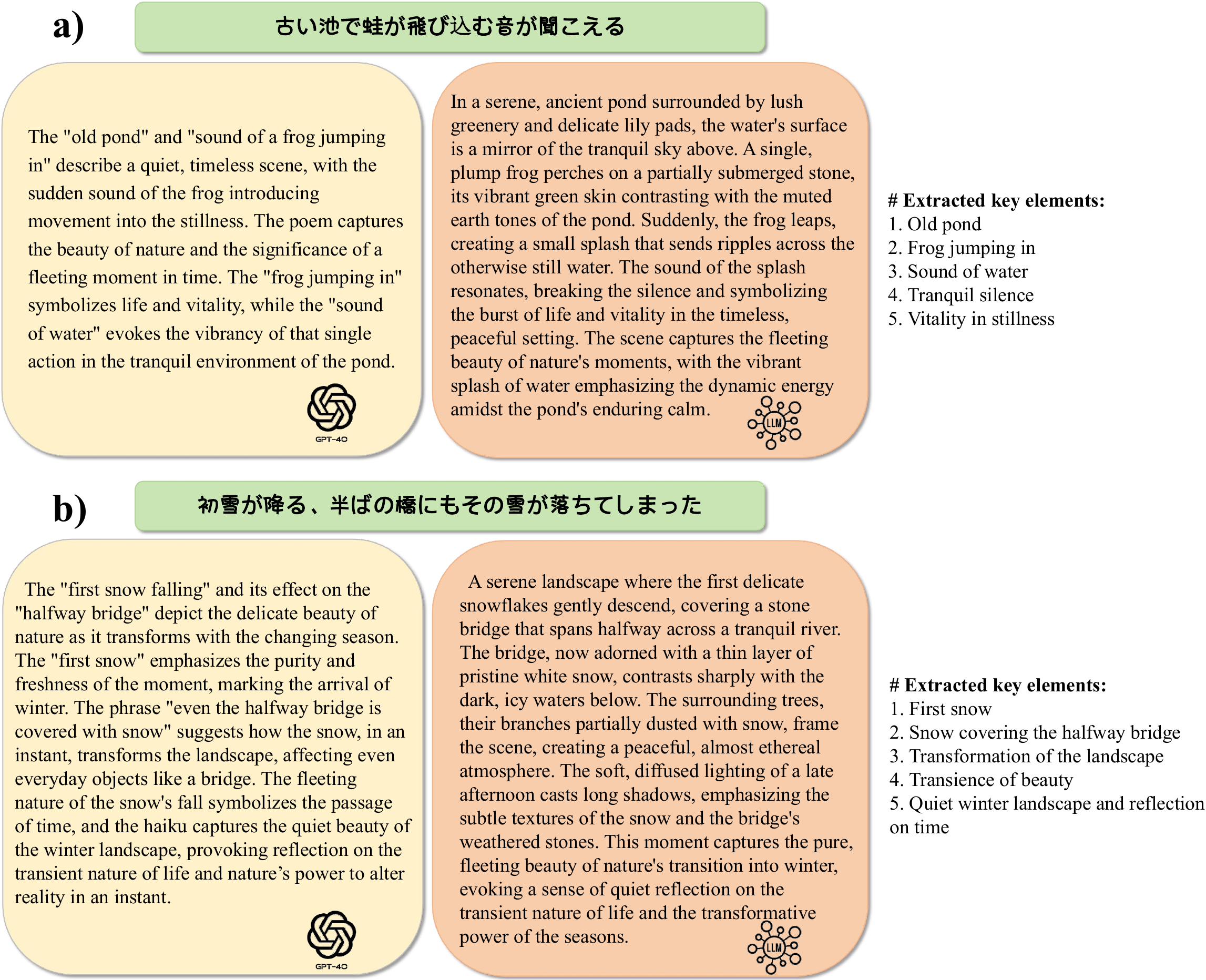}
    \caption{Haiku Enhancement \textbf{integrating traditional literary analysis principles}}
    \vspace{-2mm}
    \label{gpt1}
\end{figure*}

\begin{table}[h]
\centering
\caption{Mean Opinion Scores (1-5) from User Study}
\label{user}
\setlength{\tabcolsep}{2.0mm}
\resizebox{0.2\textwidth}{!}{
\begin{tabular}{cc}
\hline
           & \multicolumn{1}{c}{Suitable (Score 1-5)}  \\ \hline
GPT  & 3.910 $\pm$ 0.122                          \\
{LLM}        & \textbf{4.340 $\pm$ 0.111}         \\ \hline
\end{tabular}
}
\end{table} 

\begin{figure*}[h]
    \centering
    \includegraphics[width=\textwidth]{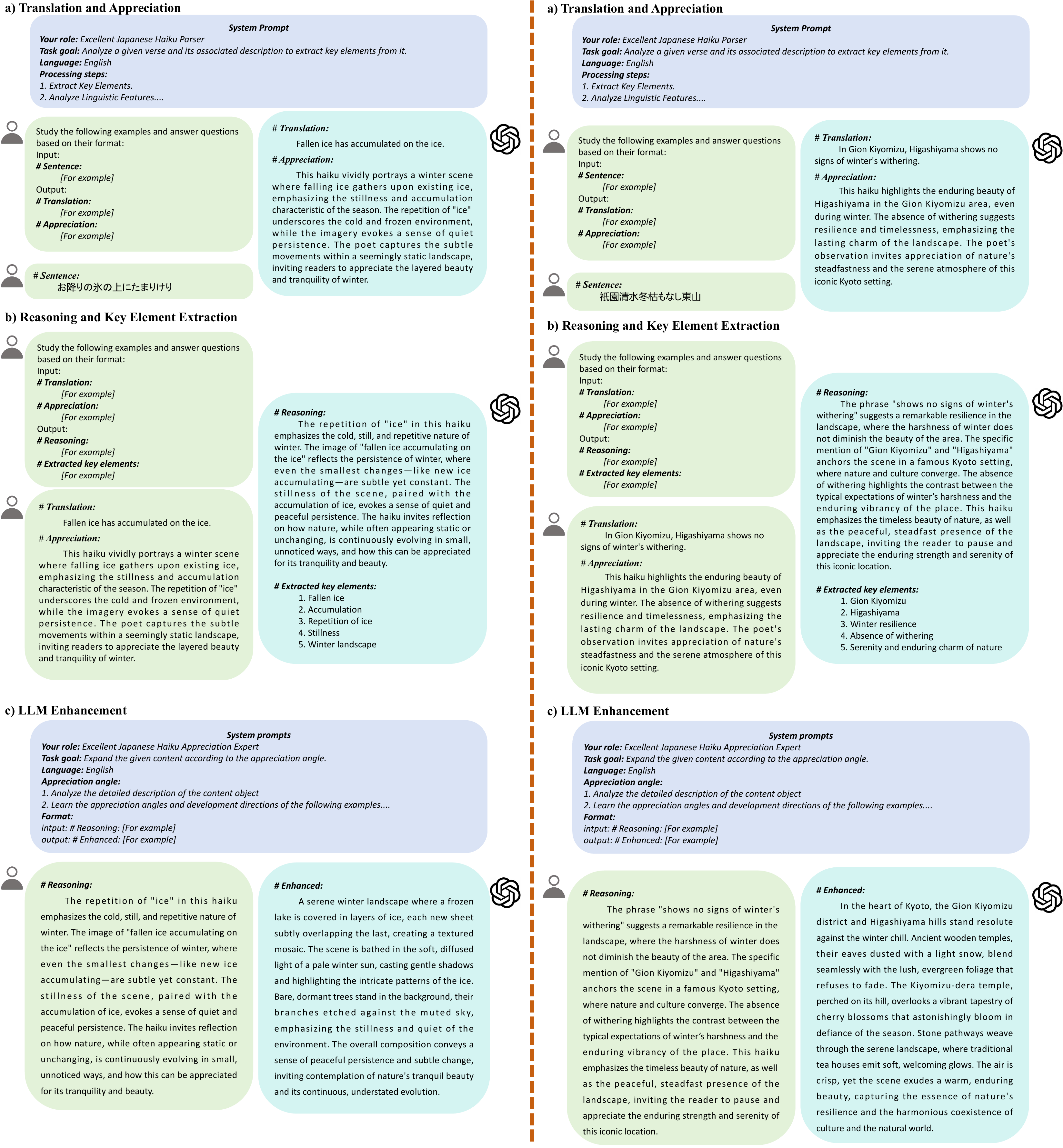}
    \caption{Haiku Enhancement \textbf{integrating traditional literary analysis principles}}
    \vspace{-2mm}
    \label{gpt2}
\end{figure*}

Additionally, we provided more examples of the text enhancement process for Haiku using LLMs, as shown in Fig. \ref{gpt1} and Fig. \ref{gpt2}.

\begin{figure*}[h]
    \centering
    \includegraphics[width=\textwidth]{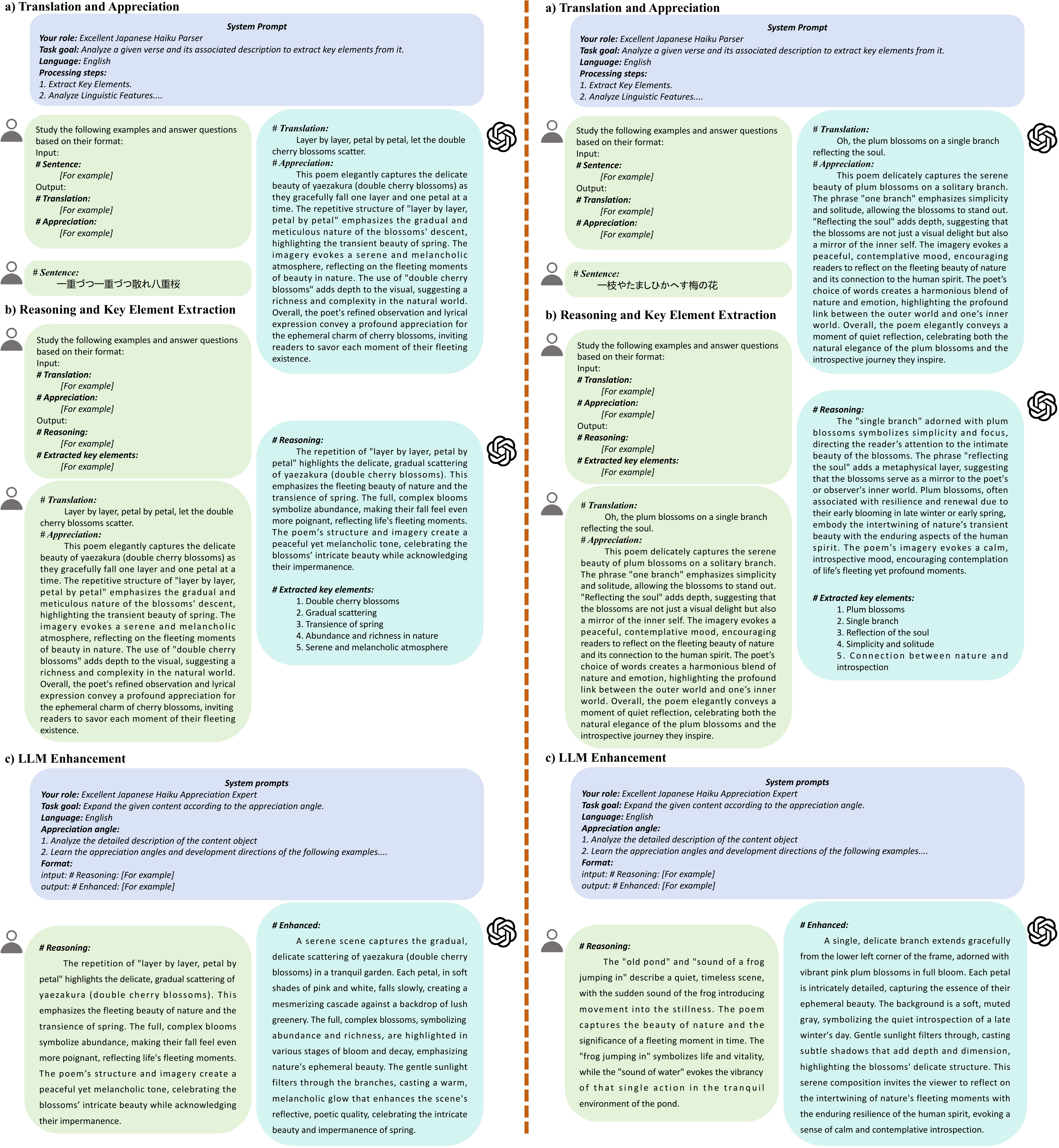}
    \caption{Haiku Enhancement \textbf{integrating traditional literary analysis principles}}
    \vspace{-2mm}
    \label{gpt1}
\end{figure*}

\bibliographystyle{named}
\bibliography{sup}